\documentclass[letterpaper, 10 pt, conference]{IEEEtran}
\IEEEoverridecommandlockouts
\usepackage[left=1.91cm,right=1.91cm,top=2.54cm,bottom=1.91cm]{geometry}

\usepackage{cite}
\usepackage{amsmath,amssymb,amsfonts}
\usepackage[caption=false,font=footnotesize]{subfig}
\usepackage{graphicx}
\usepackage{textcomp}
\usepackage{xcolor}
\usepackage{siunitx}
\usepackage{url}
\usepackage{multirow}
\usepackage{multicol}

\begin{document}

\title{Quantitative and Qualitative Assessment of Indoor Exploration Algorithms for Autonomous UAVs 
\thanks{
This work has been supported by the European Union's Horizon 2020 research and innovation programme under grant agreement No 739551 (KIOS CoE) and from the Republic of Cyprus through the Deputy Ministry of Research, Innovation and Digital Policy.}
}

\author{\IEEEauthorblockN{Adil~Farooq\IEEEauthorrefmark{1}\IEEEauthorrefmark{2}, Christos~Laoudias\IEEEauthorrefmark{1}, Panayiotis~S.~Kolios\IEEEauthorrefmark{1} and Theocharis~Theocharides\IEEEauthorrefmark{1}\IEEEauthorrefmark{2}}
\IEEEauthorblockA{\IEEEauthorrefmark{1}KIOS Center of Excellence, University of Cyprus, Nicosia, Cyprus\\
\IEEEauthorrefmark{2}Department of Electrical and Computer Engineering, University of Cyprus, Nicosia, Cyprus}
\{farooq.adil, laoudias, pkolios, ttheocharides\}@ucy.ac.cy
} 


\maketitle

\begin{abstract}

Indoor exploration is an important task in disaster relief, emergency response scenarios, and Search And Rescue (SAR) missions. Unmanned Aerial Vehicle (UAV) systems can aid first responders by maneuvering autonomously in areas inside buildings dangerous for humans to traverse, exploring the interior, and providing an accurate and reliable indoor map before the emergency response team takes action. Due to the challenging conditions in such scenarios and the inherent battery limitations and time constraints, we investigate 2D autonomous exploration strategies (e.g., based on 2D LiDAR) for mapping 3D indoor environments. First, we introduce a battery consumption model to consider the battery life aspect for the first time as a critical factor for evaluating the flight endurance of exploration strategies. Second, we perform extensive simulation experiments in diverse indoor environments using various state-of-the-art 2D LiDAR-based exploration strategies. We evaluate our findings in terms of various quantitative and qualitative performance indicators, suggesting that these strategies behave differently depending on the complexity of the environment and initial conditions, e.g., the entry point.

   
\end{abstract}



\section{Introduction}
\label{sec:intro}

In recent years, the popularity of UAVs or autonomous drones for commercial applications is quite noticeable. For instance, UAVs can be deployed for an emergency response to aid the first responders in Search And Rescue (SAR) missions~\cite{koubaa2021system}, and in collapsed buildings or structures~\cite{papaioannou2021towards}. This requires precise positioning, robustness, safety, and rapid deployment in unknown environments using only the onboard sensors, e.g., Light Detection And Ranging (LiDAR) or depth (RGB-D) camera. In 2021, such scenarios were presented as part of the DARPA Subterranean (SubT) Challenge\footnote{\url{https://www.subtchallenge.com/}} where the UAVs were operated in a highly constrained cave environment~\cite{petrlik2020robust,9476870} to perform autonomous indoor exploration missions \cite{dang2019graph}. Existing advancements in autonomous UAV exploration missions are more focused on a target-oriented approach to recognize objects of interest in the unknown environment and reach towards them efficiently~\cite{alarcon2021efficient}. The 2020 Mohamed Bin Zayed International Robotics Challenge\footnote{\url{https://www.mbzirc.com/}} (MBZIRC) showcased autonomous UAV fire-fighting exploration inside a collapsed building~ \cite{spurny2021autonomous}. Another UAV competition\footnote{\url{http://www.uasconferences.com/2022_icuas/uav-competition/}} is being organized at ICUAS 2022 that focuses on the challenges, i.e., exploration, target detection, and precise delivery faced by a fire-fighting UAV in a time-constrained urban environment. However, most of the existing solutions require a tailored UAV platform, sophisticated hardware, and are arduous to customize for real-time emergency scenarios.

UAVs are becoming capable of achieving more complex real-life SAR missions such as aiding first responders in unknown 3D environments~\cite{9213937, 9213876}. The key functionality is the ability to make intelligent decisions for navigating to targets. These decision are typically based on mobility cost, information gain, or application-specific requirements. This comes under the exploration strategies employed on a robot considering the aforementioned criteria and computation resources as a trade-off. In general, exploration algorithms operate under the assumption that the UAV is capable of gathering information and mapping the unknown environment solely through onboard sensory data. For instance, the perception sensors can be laser-based (LiDAR) or depth imaging (RGB-D) cameras. A good exploration strategy uses the collected sensor data to compute the next target effectively.

Exploration pioneering research points to seminal work done by~\cite{yamauchi1997frontier} introducing a frontier-based approach. A frontier distinguishes the boundary line between explored and unexplored regions on the map. During robot navigation the environment information gathered increases constantly, thus pushing the frontier boundary further until no frontiers are left. Such strategies have been widely tested on Unmanned Ground Vehicles (UGV)~\cite{topiwala2018frontier, 8574974, bird2021vega} in simple environments having 3~-~Degrees-of-Freedom~(DoF) i.e., ($x$, $y$, $\theta$) whereas a UAV can maneuver in more complex 3D environments with a pose of 6~-~DoF, i.e., transitional and rotational positions to control the UAV. The authors in~\cite{shen2012autonomous,selin2019efficient} proposed exploration strategies for 3D indoor environments on a UAV. This guarantees high-quality information and better performance, but at a higher cost and computational resources. To mitigate the 3D plane uncertainty in a 2D approach, it is possible to maneuver the UAV at a certain altitude above the ground plane assuming no occlusion in the Field of View (FoV) of a 2D LiDAR sensor~\cite{shen2011autonomous}. We adopt this approach and present a complete UAV system architecture in Section~\ref{sec:arch}. However, there is a limitation that 2D approaches cannot fully guarantee the occurrence of occlusions that may appear while changing the altitude.

In this work, we consider the following real-life SAR scenario that is common, yet very challenging. An emergency occurs inside a building, e.g., a fire breaks out or there is an earthquake, and the first responders need to take action to minimize human casualties. The indoor environment is partly or completely \textit{unknown} either because the indoor floorplan map is not available or, even if a map exists, it may no longer reflect the interior of the building, e.g., walls and/or ceilings are collapsing, objects are falling, etc. In this case, the first response team requires an \textit{accurate} indoor map in \textit{short time} before entering the building to operate. Thus, an autonomous UAV system with limited battery life (e.g., typically around 20-30 minutes) needs to be rapidly deployed to swiftly explore and map the unknown space under tight time constraints. UAVs using 3D exploration strategies may perform exceptionally well when there are no time limitations; however, their increased complexity and computational overhead reduces considerably the exploration time in practice leading to incomplete maps and unexplored areas. Thus, we are interested in the applicability and performance of 2D exploration strategies running on autonomous UAV systems. To this end, our contribution is twofold:

\begin{itemize}
    \item We introduce a battery consumption model that is implemented as a time-constrained battery plugin for the autonomous UAV; see Section~\ref{ssec:battery}. This plugin can reliably estimate the remaining battery life due to the UAV rotors, as well as the internal computations of various 2D exploration algorithms. To our knowledge, this is the first work to consider battery depletion in such scenarios as a critical performance indicator.
    \item We conduct an extensive comparative evaluation of three state-of-the-art 2D exploration strategies in several simulated indoor environments and analyze the experimental results with respect to various quantitative and qualitative performance indicators. We consider diverse indoor environments with varying initial conditions (e.g., different entry points for the UAV). Our findings in Section~\ref{sec:eval} indicate that there is not a clear winning solution, which motivates further research on autonomous exploration strategies for indoor SAR application scenarios.  
\end{itemize}




The rest of the paper is structured as follows. The architecture of the UAV system for autonomous exploration to aid SAR missions is detailed in Section~\ref{sec:arch}. Section~\ref{sec:explore} outlines various state-of-the-art exploration strategies and the underlying algorithms. The experimental results of our performance evaluation in diverse simulated indoor environments are presented in Section~\ref{sec:eval}. Finally, Section~\ref{sec:conc} provides concluding remarks and directions for future work.


\section{UAV System Architecture}
\label{sec:arch}

\begin{figure}[t]
 \centering
 \includegraphics[width=\linewidth]{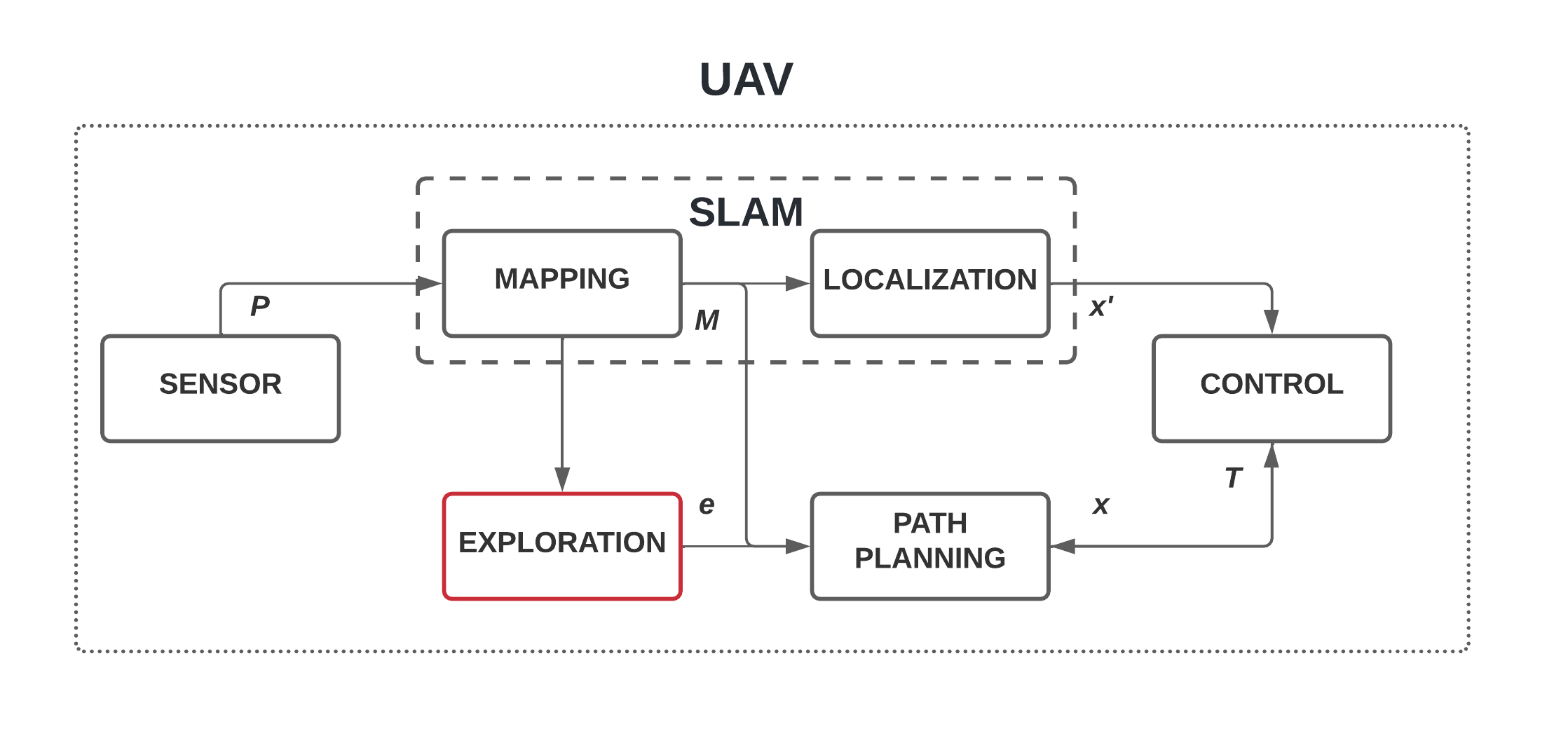} \caption{Architecture of the UAV system for SAR missions.} \label{fig:explore}
\end{figure}




There are several subsystems of a UAV system architecture to achieve full autonomy, namely \textit{Mapping}, \textit{Localization}, \textit{Path Planning}, \textit{Control}, and \textit{Exploration}, as shown in Fig.~\ref{fig:explore}. In this architecture, $P$ denotes the point cloud obtained by a 2D LiDAR, $M$ represents the map of the environment, $e$ is the current exploration node, $T$ is the planned path, $x$ is the current trajectory generated by the controller, and $x^{\prime}$ is the current pose estimate. The inner dashed lines represent the integration of two subsystems to form Simultaneous Localization and Mapping (SLAM).

Our study entails the implementation of all these subsystems using the Robot Operating System (ROS) that constitute the building foundation upon which different exploration strategies (red block in Fig.~\ref{fig:explore}) can be evaluated.


\subsection{Simultaneous Localization and Mapping}
\label{ssec:slam}

An autonomous UAV for indoor navigation in GPS-denied surroundings requires SLAM to ascertain its location in the unknown environment and gather the sensor data to construct the map. In our system architecture, we have adopted the grid-based 2D SLAM~\cite{grisetti2007improved} referred to as GMapping\footnote{http://wiki.ros.org/gmapping}, which constructs the occupancy grid or floor maps from laser scan data. It incorporates a Rao-Blackwellized Particle Filter SLAM approach to update the map, while keeping a minimum number of particles. This leads to a reduction in uncertainty of a robot’s pose, position, and orientation, on the map. Moreover, it is widely reported that GMapping is reliable for robots with inaccurate odometry and is one of the widely used 2D SLAM techniques in robotics~\cite{juneja2019comparative,chow2019toward}.

\subsection{Path Planning}
\label{ssec:planning}

A path planning problem consists of finding the optimal or the shortest path between the starting and goal positions in a static map given the robot's initial pose. This constitutes a \textit{global} and a \textit{local} planner for navigation. We have incorporated an A-star algorithm for the two exploration strategies, described in Section~\ref{ssec:wfd} and Section~\ref{ssec:explore_lite}, respectively, while a sampling-based Rapidly-exploring  Random  Tree (RRT)~\cite{8202319} was used as our global planner for the exploration strategy, described in Section~\ref{ssec:rrt}.

A local planner is used during the navigation to generate forward simulated kinematic trajectories locally on the global map to avoid static and dynamic obstacles based on the cost map parameters. For our local planner, we incorporated the Dynamic-Window Approach (DWA) proposed in~\cite{fox1997dynamic} for all the three exploration strategies.

\subsection{Control}
\label{ssec:control}

\begin{figure}[h]
 \centering
 \includegraphics[width=\linewidth]{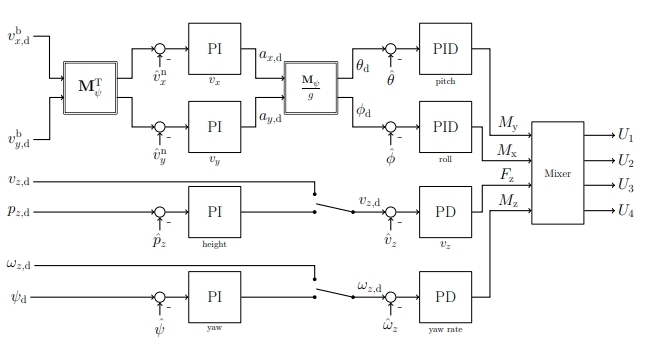} 
 \caption{Cascaded PID controllers for
roll, pitch, yaw rate and vertical velocity~\cite{meyer2012comprehensive}.} 
\label{fig:control}
\end{figure}

The control subsystem for the UAV quadcopter is realized in Fig.~\ref{fig:control} as a set of cascaded PID controllers to control the attitude, yaw, and vertical velocity in the inner loop while altitude, heading, and horizontal velocity is controlled by the outer loop of the controller. The inner loop generates the required commanded torques and thrust for all four rotors that are translated to motor voltages to control the rotor speed. The parameters of the control is set to get a desirable response during hovering and while performing complex maneuvers. 

\section{Exploration Algorithms}
\label{sec:explore}
Exploration is defined as the selection of goal points that yield the highest contribution to a specific gain function, size or closest distance from a frontier in an unknown environment. Frontiers are calculated throughout the mapping and exploration stages between known
and unknown boundary regions.
In the following, we outline state-of-the-art exploration strategies reported in the literature.

Recently, the authors in~\cite{cavinato2021dynamic} introduced a novel dynamic-aware 3D LiDAR-based exploration strategy from the seminal work on frontier exploration~\cite{yamauchi1997frontier}, improving the cost function for efficient frontier detection. However, the exploration approach was evaluated for a UGV in populated environments taking into account dynamic obstacles. In another study reported in~\cite{cieslewski2017rapid}, a frontier-based exploration strategy was proposed selecting the frontiers only in the FoV of a depth camera mounted on a UAV quadcopter to allow minimal change in velocity, in order to increase the total path length. While this approach is a good strategy to reach far unknown places, it may fail to achieve exploration coverage of the whole area, which is the motivation for our research.  

The exploration algorithms selected in our case study are applicable to a 2D LiDAR UAV system architecture, while previously have only been have evaluated for UGV~\cite{topiwala2018frontier, 8574974, bird2021vega}; hence, they have not been investigated for UAVs conclusively. To our knowledge, this is the first extensive study of exploration strategies carried out for UAV indoor SAR missions in a simulated ROS Gazebo environment.

\subsection{Wavefront Frontier Detector Exploration}
\label{ssec:wfd}

We consider the Wavefront Frontier Detector (WFD) approach proposed in~\cite{keidar2014efficient} as the first frontier-based exploration strategy. The WFD is an iterative frontier approach that performs a graph search over the previously known regions of the map, rather than taking the entire map into account. The authors attempt to perform frontier detection efficiently unlike in~\cite{yamauchi1997frontier}. During exploration, the entire map data is not scanned at each iteration, but only the explored known regions are taken into consideration. The algorithm runs two consecutive Breadth-First Searches (BFS) to detect the frontiers in the map and consider the median frontier as the next goal frontier. The WFD reduces the computation complexity, however, the full map search progress degenerates as the exploration progresses due to the backtracking nature of the algorithm. A ROS open-source implementation of the algorithm was later developed for a UGV in~\cite{topiwala2018frontier}. We employed the same strategy porting it on a UAV quadcopter model described in Section~\ref{ssec:uav}.


\subsection{Lightweight Frontier Exploration}
\label{ssec:explore_lite}

A greedy frontier-based autonomous exploration approach proposed by Hörner~\cite{Horner2016} was considered as the second exploration strategy for our indoor UAV missions. The main advantage of this algorithm compared to existing frontier techniques is its light-weightiness and the ease of integration in the ROS development environment. The nature of the generated frontier waypoints from an initial entry point is systematic for a given environment, suggesting a deterministic behavior of the algorithm. Existing works~\cite{8574974, bird2021vega} demonstrate the effectiveness of the lightweight frontier exploration on UGVs and we employed the same strategy  porting it on a UAV  quadcopter similarly to the WFD.


During the exploration search, frontiers are weighted, such that the highest weight frontier forming a centroid is sent to the navigation stack\footnote{\url{http://wiki.ros.org/navigation}} as the next frontier waypoint. The global A-star path planner then finds the shortest distance and navigates the UAV to the goal frontier. We have evaluated the approach for a single UAV exploration starting at random entry points; however, it could also be used for multi-robot and multi-session mapping~\cite{Horner2016, 7475122}.


\subsection{RRT Frontier Exploration}
\label{ssec:rrt}

The third exploration strategy proposed in~\cite{8202319} leverages Rapidly-exploring Random Trees (RRT) to detect and prioritize the frontiers in a randomized hierarchical tree structure approach. This enhances the performance from earlier described frontier-based strategies by advancing to the unknown region for frontier selection. RRT is biased towards achieving better performance in a higher order of complexity. More recently, the RRT exploration strategy for a Micro Aerial Vehicle (MAV) has been proven to be more effective than the traditional frontier-based approaches \cite{zou2022lidar}. We have evaluated the RRT approach~\cite{8202319} using the same UAV system architecture described in Section~\ref{sec:arch} for a fair comparison against the other two strategies.


\section{Simulation Setup}
\label{sec:setup}

Our UAV system was implemented in Ubuntu 18.04 Linux operating system using ROS Melodic. All simulations were performed on an Intel Core i5-8400 (6 Cores) CPU system operating at 2.80GHz with 16GB of RAM. 
\subsection{UAV platform}
\label{ssec:uav}

 Generally, a UAV quadcopter configuration is either a plus or X-shape configuration. For our experiments, we have selected the former configuration taken from~\cite{meyer2012comprehensive} providing full pose of 6-DoF, for evaluating and testing the effectiveness of all three exploration strategies. The optimal values in Table~\ref{tbl:uav} were found empirically and were kept the same in all our experiments. The UAV can maneuver within the max and min velocity range given in Table~\ref{tbl:uav} with an average velocity calculated to be \SI{0.36}{\metre/\second}. The negative sign of \SI{-0.1}{\metre/\second} shows that the UAV moves in the reverse direction to mitigate the inertia, while avoiding the obstacles.  The UAV was equipped with a 2D Hokuyo LiDAR for perception and an ultrasonic sensor for altitude control in takeoff, safe landing, and hovering~\cite{meyer2012comprehensive}. The UAV quadcopter total weight was~\SI{2}{\kilogram} with dimensions \SI{0.45}{\metre} x \SI{0.45}{\metre} x \SI{0.5}{\metre}, which lies in the small UAV category~\cite{pistoia2008battery}.

\begin{table}[th]
\caption{UAV Exploration Parameters}
\label{tbl:uav}
\centering
\begin{tabular}{|l|l|}
\hline
Name                       & value                      \\ \hline
Linear Velocity [max]  & \SI{0.5}{\metre/\second}                    \\ \hline

Linear Velocity  [min] &  \SI{-0.1}{\metre/\second}                   \\ \hline

Rotational Velocity [max]       & \SI{0.5}{\radian/\second}                  \\ \hline

Rotational Velocity [min]       & \SI{0.2}{\radian/\second}                  \\ \hline

Laser FoV                  & \SI{270}{\degree} \\ \hline
Laser Range                & \SI{8}{\metre}                        \\ \hline
Robot Clearance            & \SI{0.26}{\metre}                     \\ \hline
\end{tabular}
\end{table}

\subsection{Simulation Environment}
\label{ssec:environment}

\begin{table*}[ht]
\caption{Simulation Environment}
\label{tbl:env}
\centering
\begin{tabular}{|c|c|c|c|c|c|}
\hline
No. & Type & Size [\SI{}{\metre}]    & Area [\SI{}{\square\metre}] & Category   & Complexity     \\ \hline
$\mathcal{E}_1$   & Room     & $10\times10$ & 100  & Small      & Easy      \\ \hline
$\mathcal{E}_2$   & Apartment   & $10\times10$ & 100  & Small      & Difficult \\ \hline
$\mathcal{E}_3$   & Office   & $20\times11$ & 205  & Medium     & Moderate  \\ \hline
$\mathcal{E}_4$   & Hallway     & $20\times22$ & 342  & Medium     & Moderate  \\ \hline
$\mathcal{E}_5$   & Maze House & $20\times20$ & 400  & Large      & Difficult \\ \hline
$\mathcal{E}_6$   & School      & $70\times70$ & 4,500 & Very large & Moderate  \\ \hline
\end{tabular}
\end{table*}

\begin{figure*}[t]%
\centering
\subfloat[Room environment $\mathcal{E}_1$]{
\label{fig:E_1} 
\includegraphics[width=0.32\linewidth]{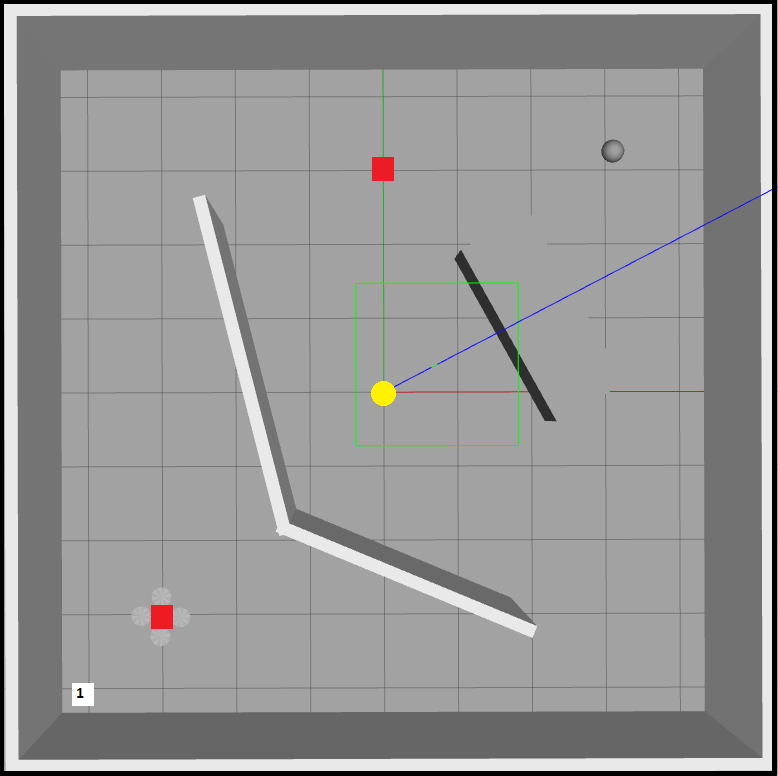}}
\hfil
\subfloat[Apartment environment $\mathcal{E}_2$]{
\label{fig:E_2} 
\includegraphics[width=0.33\linewidth]{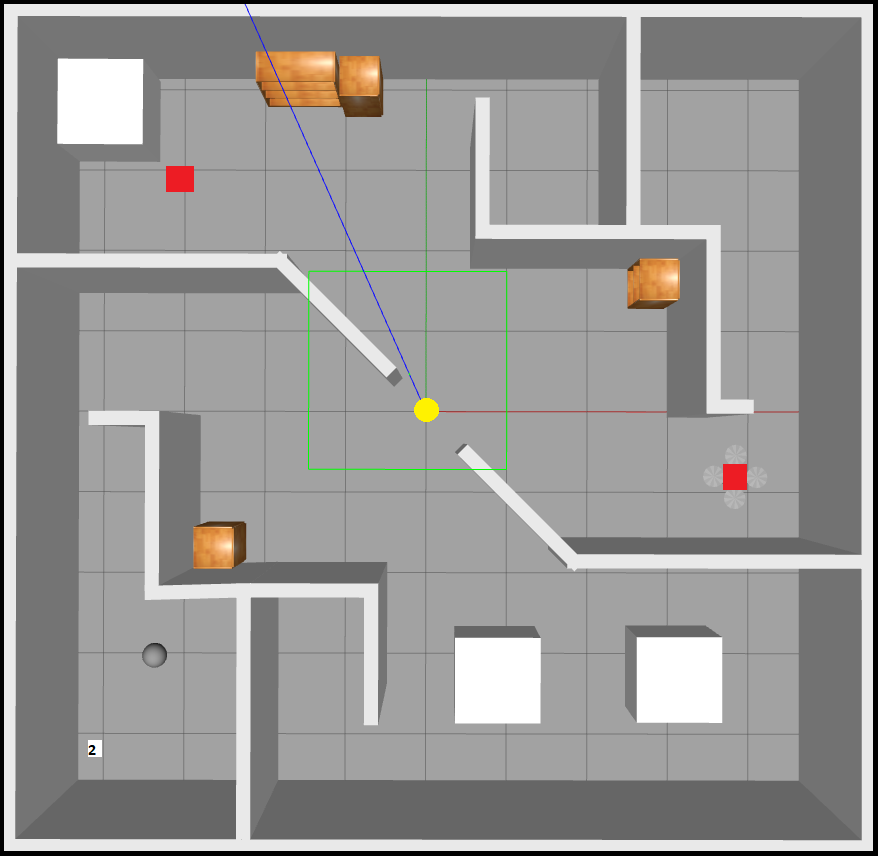}}
\hfil
\subfloat[Office environment $\mathcal{E}_3$]{
\label{fig:E_3} 
\includegraphics[width=0.31\linewidth]{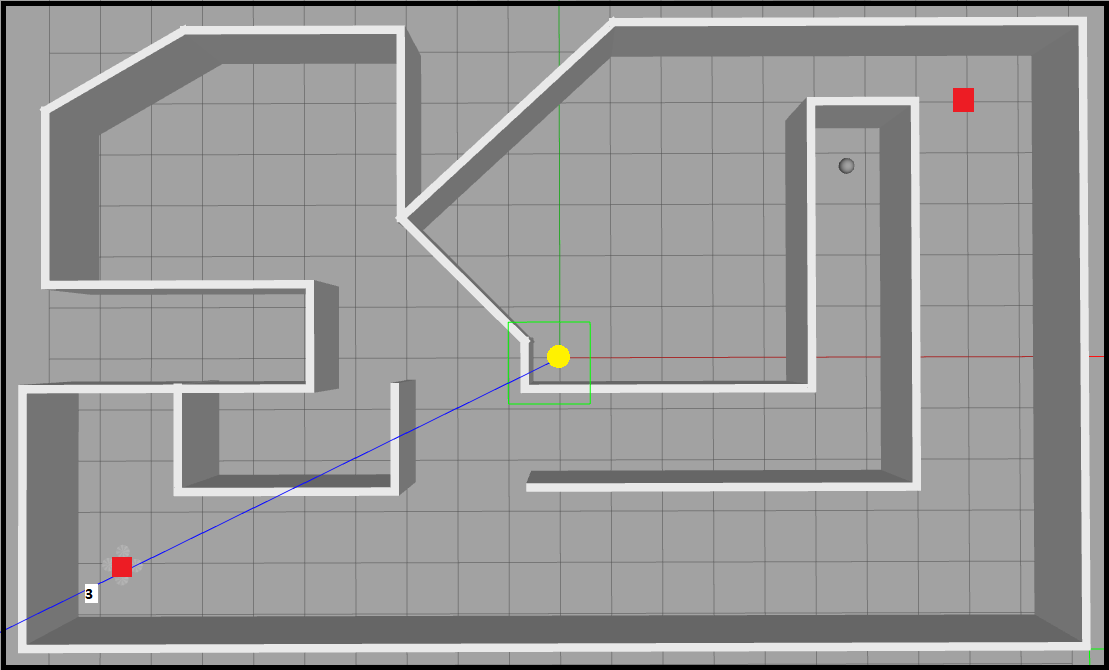}}
\hfil
\subfloat[Hallway environment $\mathcal{E}_4$]{
\label{fig:E_4} 
\includegraphics[width=0.34\linewidth]{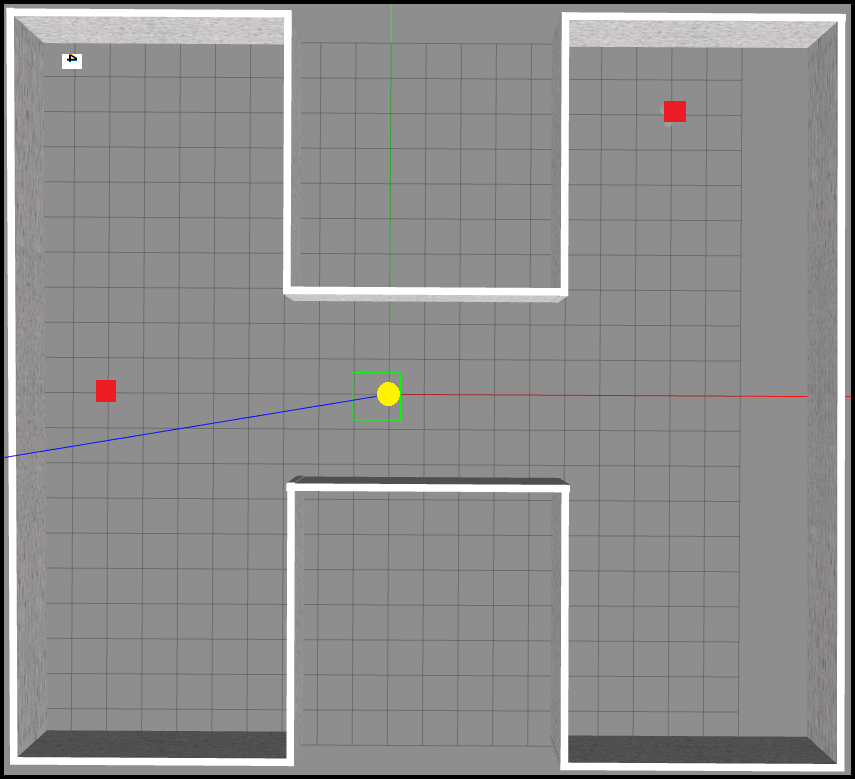}}
\hfil
\subfloat[Maze House environment $\mathcal{E}_5$]{
\label{fig:E_5} 
\includegraphics[width=0.315\linewidth]{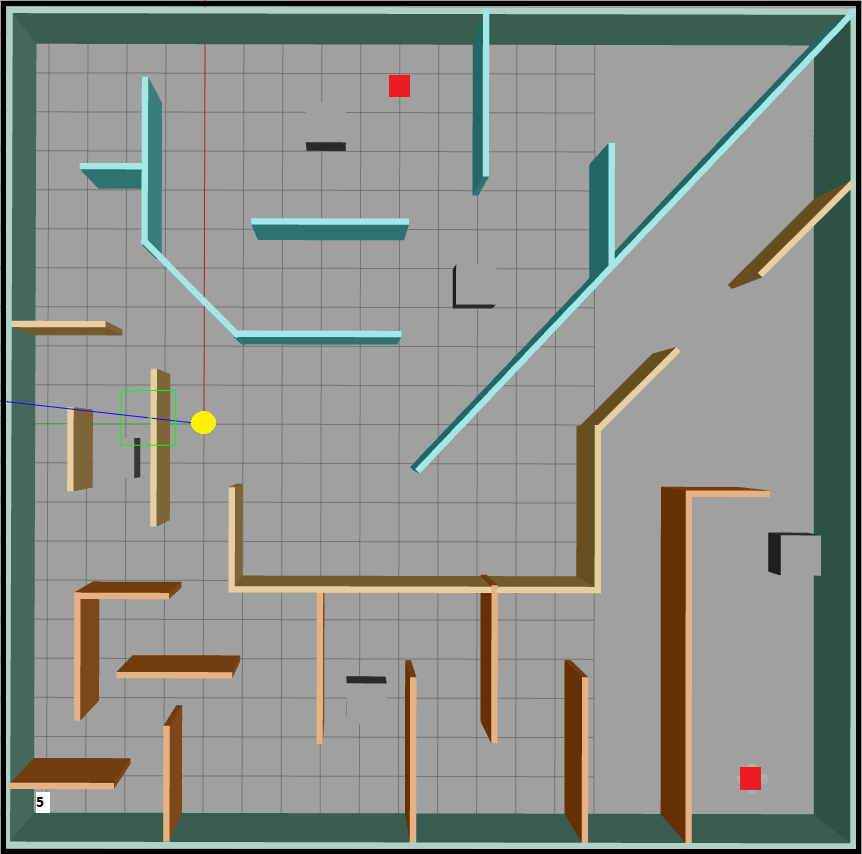}}
\hfil
\subfloat[School environment $\mathcal{E}_6$]{
\label{fig:E_6} 
\includegraphics[width=0.3\linewidth]{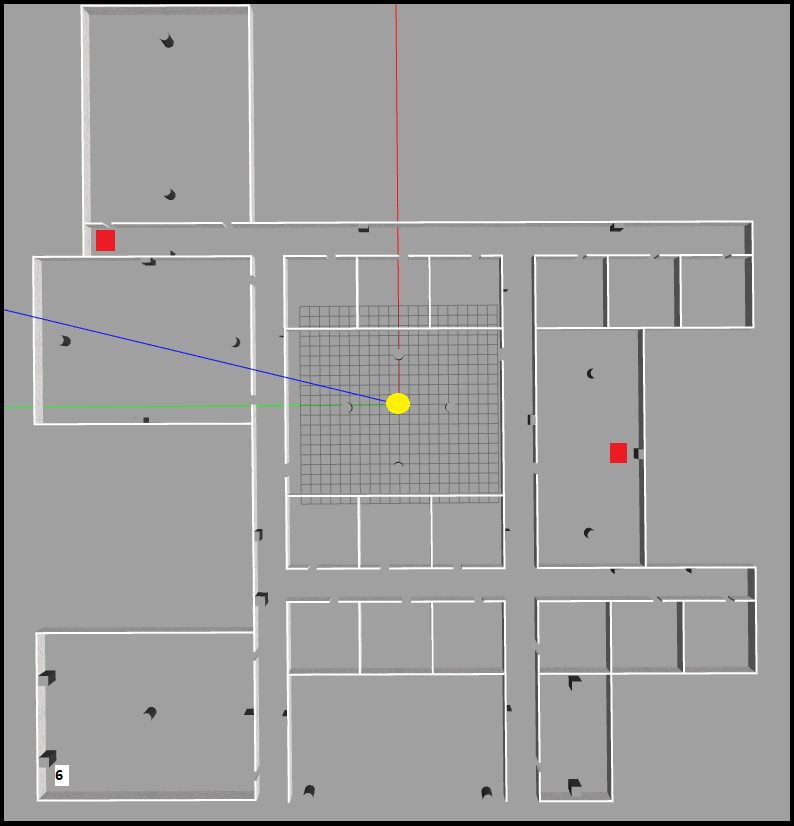}}
\caption{Exploration environments starting from entry points marked in red squares and origin (0,0) in a yellow circle.}
\label{fig:env} 
\end{figure*}


For our simulation experiments we selected six diverse and challenging indoor environments\footnote{\url{https://github.com/mferri17/ros-lidar-mapping-pathplanning}}${}^{,}$\footnote{\url{https://github.com/ethz-asl/mav_voxblox_planning}}${}^{,}$\footnote{\url{https://github.com/rumaisaabdulhai/quad_sim}} of different type and category, as well as varying size, covered area, and complexity. Specifically, they include a Room, an Apartment, an Office, a Hallway, a Maze House, and a School environment, as shown in Fig.~\ref{fig:env}. The environments are denoted $\mathcal{E}_i,~i=1,\ldots,6$ and their individual properties are listed in Table~\ref{tbl:env}. In all our simulations, the obstacle radius considered for calculating the next frontier goal was \SI{1}{\metre}, while the number of localization particles was set to the value of 30, at a constant altitude of \SI{0.3}{\metre}.





\subsection{{Battery Plugin}}
\label{ssec:battery}

We consider battery life as a critical performance indicator during exploration, thus we introduce a representative battery consumption model. The characteristics of the battery follow a linear model\footnote{https://www.mathworks.com/help/physmod/sps/powersys/ref/battery.html} given by

\begin{equation}
V = V_0 + \gamma\left( \frac{1 - i_t}{Q} \right) - Ri,
\label{eqn:B_l}
\end{equation}
where $V$ is the output voltage in volts, $V_0$ is the initial fully charged voltage in volts, $\gamma$ is the linear discharge coefficient, $i_t$ is the accumulated discharge current in ampere-hour, $Q$ is the total capacity of the battery in ampere-hour, $R$ is the total internal resistance of the battery in ohm, and $i$ is the drawn current in ampere. During battery operating mode, the charged voltage drops linearly with respect to the drawn current. A fail-safe mode~\cite{6413409} is also incorporated to land the UAV when the battery level reaches a minimum threshold of 5\%. 

In our experiments, $\gamma$ was set to 0.001 which replicates the battery timing of a commercial DJI 210 quadcopter with a flight endurance of around \SI{30}{\minute}. Moreover, we also consider the battery consumption due to the execution of the underlying algorithms in addition to the battery consumed by the motors during maneuvering of the UAV. The battery consumption varies due to internal computations to run several parallel processes in the backend of the ROS environment utilizing the CPU/GPU memory resources. Hence, the exploration time also varies depending on the nature of the explored environment. We can increase the flight endurance by mounting small lightweight solar-cell panels similar to~\cite{hassan2016solar, farooq2016modeling} on the UAV.



\section{Performance Evaluation}
\label{sec:eval}

We assess the performance of the exploration algorithms described in Section~\ref{sec:explore} with respect to a number of quantitative and qualitative Key Performance Indicators (KPI). These include the exploration cost ($E_c$), exploration time ($T_t$), exploration efficiency ($E_f$), remaining battery level ($B_l$), map completeness ($M_c$), and successful missions ($M_s$).


\subsection{Performance Indicators}
\label{ssec:metric}

For the quantitative assessment we consider $E_c$, $T_t$, $E_f$, and $B_l$,  while for the qualitative assessment we consider $M_c$ and $M_s$, which are defined in the following.

\subsubsection{Exploration Cost}
\label{sssec:Exp_c}

The exploration cost $E_c$ is defined as the total distance of the path traversed by the UAV during the exploration mission and is given by

\begin{equation}
    E_c = \sum_{i=1}^n d_i,
    \label{eqn:Exp_c}
\end{equation}
where $d_i$ is the length of the path segment $i$ in meters and $n$ is the number of piecewise linear segments that comprise the path. This indicator is similar to the multi-robot metric used in \cite{7353852} and in our case it provides a good approximation of the UAV's mobility throughout the mission.

\subsubsection{Exploration Time}
\label{sssec:Exp_t}

The exploration time $T_t$ is straightforward and is defined as the total time in minutes from the start of the mission until the mission is completed or the exploration algorithm has terminated. An effective exploration strategy minimizes $T_t$, while making the UAV maneuver to different locations that maximize the discovered area.  

\subsubsection{Exploration Efficiency}
\label{ssec:Exp_e}

The exploration efficiency $E_f$ represents the ratio of information retrieved from mapping over the total path traversed by the UAV and is computed as \cite{7353852}

\begin{equation}
    E_f = \frac{A_e}{E_c},
    \label{eqn:Exp_e}
\end{equation}
where $A_e$ is the explored area in \SI{}{\square\metre} and $E_c$ is the exploration cost given by (\ref{eqn:Exp_c}). Intuitively, the higher the $E_f$ is the more area is explored for every meter that the UAV moves. 

\subsubsection{Battery Consumption}
\label{sssec:battery_level}

Based on the battery discharge model in Section~\ref{ssec:battery}, we report the battery consumption as the \% of battery level $B_l$ that is still available either when the mission is completed (i.e., the environment has been fully explored) or the algorithm that implements the exploration strategy has terminated and no more space can be further explored.

\subsubsection{Map Completeness}
\label{sssec:Map_c}

One of the major challenges of exploration is to fully reconstruct the map of the unknown indoor environment. If this cannot be achieved, then a map that is explored to the largest possible extent is highly desirable. Assuming prior knowledge of the indoor environment and the total area it covers (i.e., ground truth), the map completeness $M_c \in [0,1]$ is given by \cite{7353852}

\begin{equation}
    M_c = \frac{A_e}{A_t},
\end{equation}
where $A_e$ and $A_t$ are the explored area and the total area in \SI{}{\square\metre}, respectively.

\subsubsection{Successful Missions}
\label{sssec:goal}

Exploration strategies do not operate in a deterministic way due to uncertainties, errors, and randomness in the underlying exploration algorithms. In addition, their behavior can be affected by initial conditions (e.g., the start point of the mission inside the indoor environment). Thus, when a mission is repeated it is often the case that the UAV does not complete the mission successfully (i.e., explore the whole area), even if the initial conditions are the same. To this end, the successful missions indicator $M_s$ provides the ratio of completed missions over the number of times that the mission is repeated. This indicator reflects the effectiveness and robustness of the exploration strategy across different indoor environments.  

\subsection{Quantitative Assessment}
\label{ssec:quantitative}

\begin{table*}[ht]
    \centering
    \caption{WFD Results}
    \label{tab:WFD_aggregate_results}
    \begin{tabular}{|c|c|c|c|c|c|c|c|}
        \hline
         Environment & Entry Point & \multicolumn{6}{c|}{WFD Frontier} \\
         \cline{3-8} 
         No. & (x,y) & $E_c$ [\SI{}{\metre}] & $B_l$ [\%] & $T_t$ [\SI{}{\minute}] & $E_f$ & $M_c$ & $M_s$\\
         \hline
         
         \multirow{2}{*}{$\mathcal{E}_1$} & (-3,-3) 
         & 8.4$\pm$0.6 & 96.2$\pm$1.1 & 0.5$\pm$0.3 & 11.7$\pm$0.9 & 1 & 5/5 \\
         \cline{2-8}
         \multirow{2}{*}{} & (0,3) 
         & \textbf{10.9$\pm$1.5} & 90.3$\pm$1 & 1.7$\pm$0.3 & \textbf{9.3$\pm$1.1} & 1 & 5/5 \\
         \cline{1-8}
         
         \multirow{2}{*}{$\mathcal{E}_2$} & (3.8,-0.8) 
         & 41.2$\pm$6.2 & \textbf{65.6$\pm$1.3} & \textbf{7.4$\pm$0.4} & 2.3$\pm$0.3 & 0.9 & 0/5\\
         \cline{2-8}
         \multirow{2}{*}{} & (-2.5,2.5) 
         & 31.5$\pm$12.2 & \textbf{81.1$\pm$8.8} & \textbf{3.8$\pm$2.0} & \textbf{3.3$\pm$1.6} & 0.9 & 0/5\\
         \cline{1-8}
         
         \multirow{2}{*}{$\mathcal{E}_3$} & (-8.5,-4) 
         & 56.2$\pm$22.9 & 62.5$\pm$6.6 & 7.7$\pm$1.2 & 3.6$\pm$1.3 & 0.9$\pm$0.1 & 0/5\\
         \cline{2-8}
         \multirow{2}{*}{} & (8,5) 
         & 58.4$\pm$9.3 & 49.6$\pm$9.4 & 10.6$\pm$2.1 & 3.4$\pm$0.6 & 0.9 & 0/5\\
         \cline{1-8}
         
         \multirow{2}{*}{$\mathcal{E}_4$} & (8,8) 
         & 37$\pm$5.6 & 73.6$\pm$2 & 4.8$\pm$0.4 & 9.4$\pm$1.4 & 1 & 5/5\\
         \cline{2-8}
         \multirow{2}{*}{} & (-8,0) 
         & 32.8$\pm$5.7 & 86.1$\pm$2.4 & 2.5$\pm$0.6 & 10.7$\pm$1.9 & 1 & 5/5\\
         \cline{1-8}
         
         \multirow{2}{*}{$\mathcal{E}_5$} & (-9,14) 
         & \textbf{139$\pm$17.9} & 10.5$\pm$7.3 & 14.9$\pm$1.5 & \textbf{2.9$\pm$0.4} & \textbf{0.9$\pm$0.1} & \textbf{1/5}\\
         \cline{2-8}
         \multirow{2}{*}{} & (8,-2) 
         & \textbf{123.2$\pm$13.4} & \textbf{19.5$\pm$13.8} & \textbf{13.2$\pm$2.3} & \textbf{3.1$\pm$0.3} & \textbf{0.9$\pm$0.1} & \textbf{1/5}\\
         \cline{1-8}
         
         \multirow{2}{*}{$\mathcal{E}_6$} & (-5,-22) 
         & 155$\pm$42.3 & 5 & 22.2$\pm$0.1 & 16.4$\pm$1.8 & 0.6 & 0/2\\
         \cline{2-8}
         \multirow{2}{*}{} & (16.5,30) 
         & 152.3$\pm$32.5 & 5 & 23.1$\pm$ & 15.7$\pm$2.4 & 0.5 & 0/2\\
         \cline{1-8}

    \end{tabular}

\end{table*}

\begin{table*}[ht]
    \centering
    \caption{Lightweight Frontier Results}
    \label{tab:lite_aggregate_results}
    \begin{tabular}{|c|c|c|c|c|c|c|c|}
        \hline
         Environment & Entry Point & \multicolumn{6}{c|}{Lightweight Frontier} \\
         \cline{3-8} 
         No. & (x,y) & $E_c$ [\SI{}{\metre}] & $B_l$ [\%] & $T_t$ [\SI{}{\minute}] & $E_f$ & $M_c$ & $M_s$\\
         \hline
         
         \multirow{2}{*}{$\mathcal{E}_1$} & (-3,-3)  
         & \textbf{8.2$\pm$0.6} & \textbf{96.3$\pm$1.6} & 0.5$\pm$0.3 & \textbf{12.3$\pm$0.9} & 1 & 5/5\\
         \cline{2-8}
         \multirow{2}{*}{} & (0,3)  
         & 11.6$\pm$1.1 & \textbf{91.1$\pm$0.9} & \textbf{1.6$\pm$0.2} & 8.6$\pm$0.8 & 1 & 5/5\\
         \cline{1-8}
         
         \multirow{2}{*}{$\mathcal{E}_2$} & (3.8,-0.8) 
         & \textbf{31.6$\pm$10.1} & 66.3$\pm$7.2 & 7.5$\pm$1.6 & \textbf{2.9$\pm$1.1} & 0.9 & 0/5\\
         \cline{2-8}
         \multirow{2}{*}{} & (-2.5,2.5) 
         & \textbf{31.9$\pm$6.3} & 79.7$\pm$7.1 & 4$\pm$1.4 & 2.9$\pm$0.5 & 0.9 & 0/5\\
         \cline{1-8}
         
        \multirow{2}{*}{$\mathcal{E}_3$} & (-8.5,-4)  
        & \textbf{45.5$\pm$21.7} & \textbf{67.2$\pm$8.1} & \textbf{6.8$\pm$1.7} & \textbf{4.4$\pm$1.6} & 0.7$\pm$0.1 & 0/5\\
         \cline{2-8}
         \multirow{2}{*}{} & (8,5) 
         & \textbf{55.6$\pm$10.1} & \textbf{50.9$\pm$10.6} & \textbf{10.1$\pm$3.1} & \textbf{3.4$\pm$0.7} & 0.8 & 0/5\\
         \cline{1-8}
         
         \multirow{2}{*}{$\mathcal{E}_4$} & (8,8) 
         & 31.9$\pm$6.1 & \textbf{75$\pm$3.6} & \textbf{4.4$\pm$0.7} & 11$\pm$2.1 & 1 & 5/5\\
         \cline{2-8}
         \multirow{2}{*}{} & (-8,0) 
         & \textbf{21.5$\pm$2} & \textbf{90.9$\pm$1.1} & \textbf{1.5$\pm$0.3} & \textbf{15.9$\pm$1.5} & 1 & 5/5\\
         \cline{1-8}
         
         \multirow{2}{*}{$\mathcal{E}_5$} & (-9,14) 
         & 149.4$\pm$16 & \textbf{12.6$\pm$10.4} & \textbf{14.7$\pm$1.8} & 2.4$\pm$0.3 & 0.9$\pm$0.1 & 0/5\\
         \cline{2-8}
         \multirow{2}{*}{} & (8,-2) 
         & 130.9$\pm$8.2 & 10.6$\pm$3.9 & 14.7$\pm$0.6 & 2.9$\pm$0.1 & 0.9 & 0/5\\
         \cline{1-8}
         
         \multirow{2}{*}{$\mathcal{E}_6$} & (-5,-22) 
         & 143.1$\pm$27.7 & 5 & 22.2$\pm$1 & 16.2$\pm$1.1 & 0.6 & 0/2\\
         \cline{2-8}
         \multirow{2}{*}{} & (16.5,30) 
         & 140.3$\pm$19.3 & 5 & 22.7$\pm$1.7 & 15.8$\pm$2.3 & 0.5 & 0/2\\
         \cline{1-8}

    \end{tabular}

\end{table*}

\begin{table*}[ht]
    \centering
    \caption{RRT Frontier Results}
    \label{tab:rrt_aggregate_results}
    \begin{tabular}{|c|c|c|c|c|c|c|c|}
        \hline
         Environment & Entry Point & \multicolumn{6}{c|}{RRT Frontier} \\
         \cline{3-8} 
         No. & (x,y) & $E_c$ [\SI{}{\metre}] & $B_l$ [\%] & $T_t$ [\SI{}{\minute}] & $E_f$ & $M_c$ & $M_s$\\
         \hline
         
         \multirow{2}{*}{$\mathcal{E}_1$} & (-3,-3) 
         & 11.4$\pm$1.4 & 93.7$\pm$1.3 & 1.3$\pm$0.1 & 8.8$\pm$1 & 1 & 5/5\\
         \cline{2-8}
         \multirow{2}{*}{} & (0,3) 
         & $-$ & $-$ & $-$ & $-$ & $-$ & 0/5\\
         \cline{1-8}
         
         \multirow{2}{*}{$\mathcal{E}_2$} & (3.8,-0.8) 
         & $-$ & $-$ & $-$ & $-$ & $-$ & 0/5\\
         \cline{2-8}
         \multirow{2}{*}{} & (-2.5,2.5) 
         & 37.8$\pm$13.8 & 76.1$\pm$8.5 & 5.9$\pm$2.4 & 3$\pm$1.3 & 0.9 & 0/5\\
         \cline{1-8}
         
        \multirow{2}{*}{$\mathcal{E}_3$} & (-8.5,-4)  
        & 81.2$\pm$16.8 & 48.7$\pm$11.5 & 12.9$\pm$3.2 & 2.6$\pm$0.5 & \textbf{1} & \textbf{5/5}\\
         \cline{2-8}
         \multirow{2}{*}{} & (8,5) 
         &  $-$ &  $-$ &  $-$ &  $-$ & $-$ & 0/5\\
         \cline{1-8}
         
         \multirow{2}{*}{$\mathcal{E}_4$} & (8,8) 
         & \textbf{25.3$\pm$9.5} & 66.4$\pm$9.2 & 8$\pm$2.8 & \textbf{12.5$\pm$2.1} & 0.9$\pm$0.3 & 4/5\\
         \cline{2-8}
         \multirow{2}{*}{} & (-8,0) 
         & 32.5$\pm$3.9 & 85.8$\pm$1.3 & 2.9$\pm$0.4 & 10.6$\pm$1.3 & 1 & 5/5\\
         \cline{1-8}
         
         \multirow{2}{*}{$\mathcal{E}_5$} & (-9,14) 
         & 160.2$\pm$27.1 & 7.4$\pm$5.3 & 15.4$\pm$1.0 & 2.3$\pm$0.3 & 0.9 & 0/5\\
         \cline{2-8}
         \multirow{2}{*}{} & (8,-2) 
         & 111.3$\pm$40.2 & 15.8$\pm$15.5 & 13.6$\pm$2.9 & 2.4$\pm$0.6 & 0.6$\pm$0.1 & 0/5\\
         \cline{1-8}
         
         \multirow{2}{*}{$\mathcal{E}_6$} & (-5,-22) 
         & \textbf{179.1$\pm$11.5} & 5 & \textbf{29.3$\pm$5.8} & \textbf{17.3$\pm$0.4} & \textbf{0.7} & 0/2\\
         \cline{2-8}
         \multirow{2}{*}{} & (16.5,30)
         & \textbf{159.5$\pm$15.8} & 5 & \textbf{38.3$\pm$1.1} & \textbf{19.1$\pm$1} & \textbf{0.7} & 0/2\\
         \cline{1-8}

    \end{tabular}

\end{table*}

In our study, we conducted 54 simulation experiments in each of the six simulated indoor environments and for each exploration strategy. In total 162 simulation experiments were performed for the three exploration algorithms. For all the runs, the cutoff value used in clustering a frontier goal was set to \SI{0.5}{\metre}. The aggregated results in terms of mean and standard deviation were tabulated in Table~\ref{tab:WFD_aggregate_results}, Table~\ref{tab:lite_aggregate_results}, and Table~\ref{tab:rrt_aggregate_results}, respectively.

For the quantitative assessment, we performed experiments in all the environments listed in Table~\ref{tbl:env}. The results were compared in terms of the quantitative KPIs explained in Sub-Section~\ref{ssec:metric}, i.e, $E_c$, $B_l$, $T_t$, and $E_f$. The best results are shown for the selected KPIs in Table~\ref{tab:WFD_aggregate_results}, Table~\ref{tab:lite_aggregate_results}, and Table~\ref{tab:rrt_aggregate_results}, respectively, will be compared both in terms of numeric values highlighted in bold and from an algorithmic perspective. However, for brevity, we will only give insights into some of the evaluated environments, i.e., for a small apartment $\mathcal{E}_2$, an office $\mathcal{E}_3$, and a very spacious school $\mathcal{E}_6$ environment while the remaining environments can be interpreted in the same manner.  


In case of $\mathcal{E}_2$, shown in Fig.~\ref{fig:E_2}, we considered the two entry points at (3.8,-0.8) and (-2.5,2.5), respectively. It has several narrow passages with obstacles. The lightweight frontier approach showed better performance in terms of $E_c=\SI{31.6}{\metre}$, $E_f=2.9$ while WFD was better in terms of $B_l=\SI{65.6}{\%}$ with a small standard deviation of $\pm$1.3, compared to $B_l=\SI{66.3}{\%}$ with much higher standard deviation of  $\pm$7.2 for lightweight frontier approach, and $T_t=\SI{7.4}{\minute}$. In contrary, the RRT exploration failed to start after multiple trials, even taking different initial bias points. For the second entry point, the lightweight frontier approach $E_c=\SI{31.9}{\metre}$ showed small standard deviation of $\pm$6.3 compared to WFD $E_c=\SI{31.5}{\metre}$ $\pm$12.2. However, the WFD approach performed better in terms of the three indicators, i.e., $B_l=\SI{81.1}{\%}$, $T_t=\SI{3.8}{\minute}$, and $E_f=3.3$, respectively. The higher variation in $E_c$ for WFD is because, in each trial, the algorithm searches for the median frontier unlike finding a centroid in the case of lightweight frontier approach. The RRT exploration approach again proved to be ineffective in a small and congested indoor environment due to its high dimensional algorithmic nature.   

We considered for evaluation a medium sized environment $\mathcal{E}_3$, shown in Fig.~\ref{fig:E_3}, with a covered area of \SI{205}{\square\metre} less than its total area, due to its diverse and irregular structure. For the first entry point at (-8.5,-4), we see that lightweight frontier had the optimal performance in terms of all the four quantitative KPIs compared to the other two approaches. Similarly, for the second entry point at (-8.5,-4), lightweight approach is the optimal choice, while RRT failed to start from this position. Hence, the lightweight frontier approach is the best option in $\mathcal{E}_3$. Notably, the RRT exploration was the only strategy to successfully complete the mission in each trial when starting from the first entry point, making it a more robust and reliable approach from it algorithmic perspective.

Lastly, we consider the environment $\mathcal{E}_6$ to be a very large spacious school, shown in Fig.~\ref{fig:E_6}. Here, we assume the higher values of the $E_c$, $T_t$ and $E_f$ KPIs will be the more the exploration coverage is achieved. This is contrary to our earlier assumption made in the previous five environments where the battery consumption was taken into account as a \textit{critical factor}. Whenever the battery depletes to a minimum level, in our case it is set to $\SI{5}{\%}$, the UAV lands safely. It is also possible to return to the initial position when the battery depletes to a certain level during exploration as a future goal. Our purpose was to provide an evaluation of all three strategies when reaching the minimum battery threshold. We observed for the first entry point at (-5,-22), the RRT frontier outperformed in $E_c=\SI{179.1}{\metre}$, $T_t=\SI{29.3}{\minute}$, and  $E_f=17.3$. Similarly, for the second entry point at (16.5,30) the RRT frontier achieved the best results $E_c=\SI{159.5}{\metre}$, $T_t=\SI{38.3}{\minute}$, and  $E_f=19.1$. Essentially, the RRT covered the maximum exploration area with the lowest computational overhead compared to the former two approaches. This was expected, as the sampling-based RRT approach is biased towards high performance in larger and spacious environments.       


\subsection{Qualitative Assessment}
\label{ssec:qualitative}

\begin{figure*}[t]%
\centering
\subfloat[]{
\label{fig:spider:a} 
\includegraphics[width=0.49\linewidth]{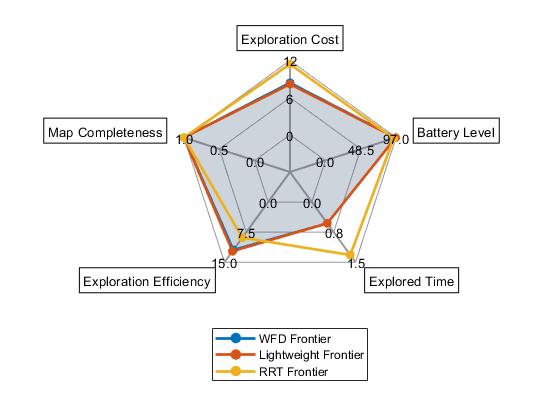}}
\hfil
\subfloat[]{
\label{fig:spider:b} 
\includegraphics[width=0.49\linewidth]{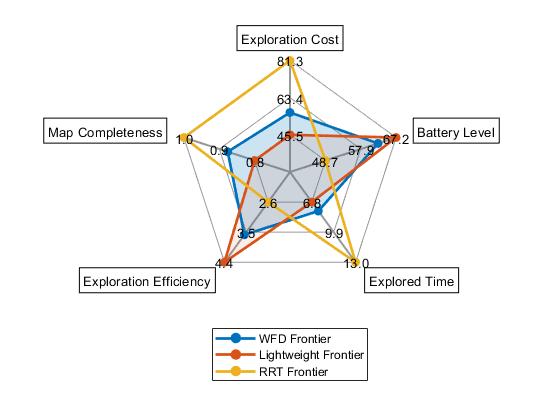}}
\hfil
\hspace{1in}
\subfloat[]{
\label{fig:spider:c} 
\includegraphics[width=0.49\linewidth]{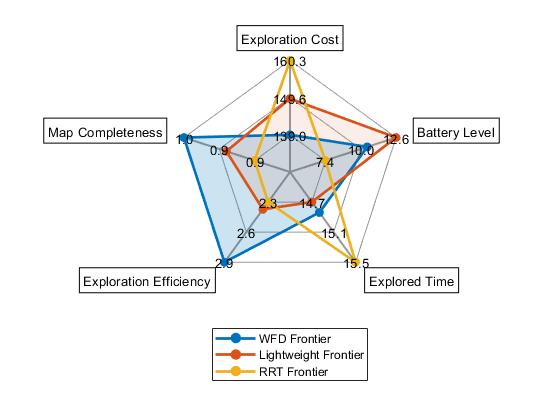}}

\caption{Exploration of indoor environment; (a) $\mathcal{E}_1$ (small), (b) $\mathcal{E}_3$ (medium) and (c) $\mathcal{E}_5$ (large).}
\label{fig:spider_plots} 
\end{figure*}

The constructed 2D occupancy/floor map provides a good qualitative measure of the exploration mission's success. The $M_c$ and $M_s$ performance indicators are evaluated qualitatively from the spider plots in Fig.~\ref{fig:spider_plots}. It is to note that the mission success is taken as a qualitative indicator and not a quantitative, as we would prefer a robust and highly accurate exploration strategy with a 100$\%$ success rate for an indoor SAR mission, instead of an incomplete mission. While for brevity, we present only the final maps constructed by all three exploration strategies for an Office $\mathcal{E}_3$ and the exploration coverage using the ROS Gazebo visualizing tool Rviz for $\mathcal{E}_6$ environment. We also provide a video demonstration of our work in the Maze House $\mathcal{E}_5$ environment\footnote{\url{https://youtu.be/moDyXkZQw0s}}.

\begin{figure*}[t]%
\centering
\subfloat[WFD exploration]{
\label{fig:indoor_3:a} 
\includegraphics[width=0.31\linewidth]{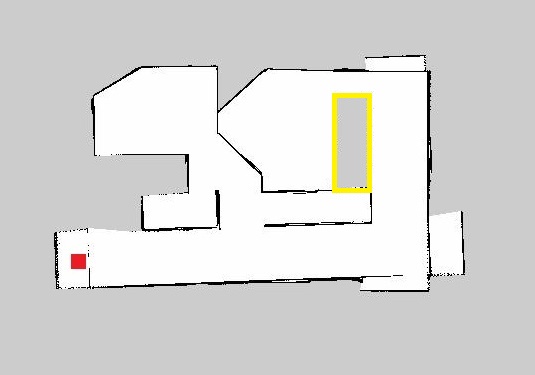}}
\hfil
\subfloat[Light-weight Frontier exploration]{
\label{fig:indoor_3:b} 
\includegraphics[width=0.31\linewidth]{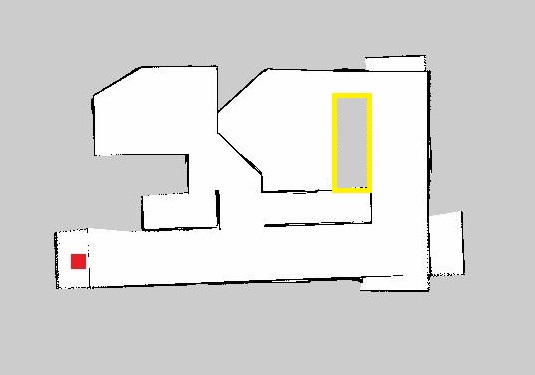}}
\hfil
\subfloat[RRT Frontier exploration]{
\label{fig:indoor_3:c} 
\includegraphics[width=0.33\linewidth]{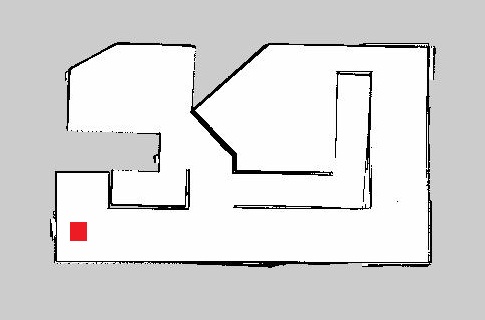}}
\caption{Exploration of the Office environment $\mathcal{E}_3$ starting from the entry point (-8.5,-4) marked in red square.}
\label{fig:indoor_3} 
\end{figure*}

\begin{figure*}[t]%
\centering
\subfloat[WFD exploration]{
\label{fig:school:a} 
\includegraphics[width=0.32\linewidth]{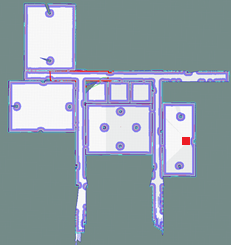}}
\hfil
\subfloat[Light-weight Frontier exploration]{
\label{fig:school:b} 
\includegraphics[width=0.32\linewidth]{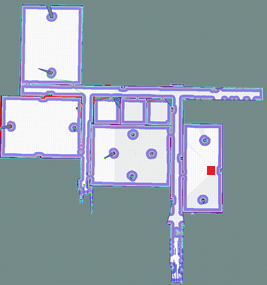}}
\hfil
\subfloat[RRT Frontier exploration]{
\label{fig:school:c} 
\includegraphics[width=0.31\linewidth]{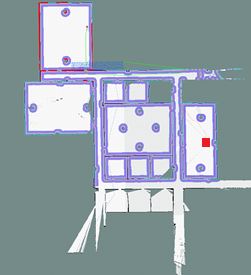}}
\caption{Exploration of the School environment $\mathcal{E}_6$ starting from the entry point (-5, -22) marked in red square.}
\label{fig:school} 
\end{figure*}

First, we selected the simulated office environment $\mathcal{E}_3$, shown in Fig.~\ref{fig:E_3}, due to its diverse structure. The UAV started its exploration from the entry point (-8.5,-4). It proves difficult for the WFD and lightweight frontier approaches to fully explore the environment due to the deterministic nature of the underlying BFS frontier detection algorithm, as shown in the constructed floor map, highlighted in a yellow rectangle box in Fig.~\ref{fig:indoor_3:a} and Fig.~\ref{fig:indoor_3:b}, respectively. This implies that both strategies were not able to find the remaining frontiers, that were not appearing in the FoV of the LiDAR. In contrary, the RRT frontier detection approach was able to completely explore and map the environment in all trials starting from the same entry point (i.e., $M_s=5/5$), as shown in Fig.~\ref{fig:spider:b}, for a medium-sized indoor office environment ($\mathcal{E}_3$). This is because the sampling-based RRT approach was able to reach the unexplored regions efficiently compared to any of the other two backtracking BFS frontier approaches. Hence, the RRT approach outperforms frontier-based approaches in diverse, cluttered and unstructured medium indoor environments.



The second environment selected for our qualitative assessment is a school $\mathcal{E}_6$ featuring multiple small and large rooms with open space, shown in Fig.~\ref{fig:E_6}. The UAV started its exploration from the entry point (-5,-22) constructed the floor map, shown in Fig.~\ref{fig:school}, where the colors, \textit{cyan} represents the UAV clearance distance and \textit{purple} obstacle avoidance radius given in Section~\ref{ssec:uav}. Here, we show the results qualitatively in terms of exploration, while the occupancy/floor map is being constructed. We observe that none of the three exploration algorithms was able to complete the mission successfully within the $95\%$ battery limit. However, we see that the exploration coverage area of RRT is larger than both WFD and lightweight frontier exploration approaches. This is because RRT is a random sampling algorithm to detect new frontiers and is biased towards finding goal frontiers in far unknown regions, while the other algorithms follow a systematic approach and are prone to backtracking described earlier.   


From the spider plots in Fig.~\ref{fig:spider_plots}, we see that the lightweight approach, shown in Fig.~\ref{fig:spider:a}, performed slightly better in terms of $M_c$ than WFD for a small room environment ($\mathcal{E}_3$), whereas WFD performs better than the lightweight approach in a large indoor environment ($\mathcal{E}_5$), as shown in Fig.~\ref{fig:spider:c}.

\subsection{Discussion}
\label{ssec:discussion}

In our experiments, we investigate in practice the performance of three state-of-the-art exploration strategies for autonomous indoor UAV missions both quantitatively and qualitatively. We consider  six KPIs i.e, $E_c$, $B_l$, $T_t$, $E_f$, $M_c$ and $M_s$ that are critical in SAR missions. The outstanding quantitative results for the first four KPIs are tabulated as bold for each exploration strategy in Table~\ref{tab:WFD_aggregate_results}, Table~\ref{tab:lite_aggregate_results} and Table~\ref{tab:rrt_aggregate_results}, respectively. In a nutshell, in small and medium-size environments, the WFD and the lightweight frontier-based strategies in most of the cases performed equally well; however, the latter proved to be more efficient in terms of all the four quantitative KPIs. In contrast, the RRT frontier strategy proved to be robust and highly efficient as the size and complexity of the environment increases, from medium, large to very large indoor environments. Thus, showing the robustness and effectiveness of sampling-based exploration strategy for such indoor settings. 

The qualitative $M_c$ and $M_s$ results were obtained from the spider plots in Fig.~\ref{fig:spider_plots} as well as from the constructed occupancy/floor maps shown in Fig.~\ref{fig:indoor_3}, and the exploration map of Fig.~\ref{fig:school}. While among the frontier-based strategies, the WFD proved to be a better choice both in terms of the  $M_c$ and $M_s$, especially in the case of $\mathcal{E}_5$ shown in Fig.~\ref{fig:spider:c}, the most challenging environment, however, it is not always the case that a successful mission is guaranteed. Like in the quantitative assessment, the RRT frontier approach is more robust and highly efficient both in terms of $M_c$ and $M_s$, as the size and complexity of the environment increases, from medium, large to very large indoor environments.

Therefore, the results in Section~\ref{ssec:quantitative} and Section~\ref{ssec:qualitative} indicate that there is no clear winning strategy for all indoor environments. With regards to the internal computational resources, such as CPU/GPU memory, the RRT frontier approach in spacious environments becomes more computationally efficient as compared to smaller enclosed congested and narrow passage environments, making sampling-based approach an ideal choice in terms of $E_c$, $B_l$ and  $M_c$ for larger indoor missions. Moreover, it can increase the remaining $B_l$ and $T_t$ significantly, as shown for $\mathcal{E}_6$ in Table~\ref{tab:rrt_aggregate_results}, thus offering additional time for further exploration. Clearly, these observations motivate further research on autonomous exploration strategies for indoor SAR application scenarios.






%
\section{Conclusions}
\label{sec:conc}

In this work, we have introduced a new KPI for exploration, i.e., battery consumption based on a battery discharging model to ensure reliable flight endurance, and critically evaluating the performance of the three exploration strategies. We performed an extensive simulation in various diverse and challenging indoor test environments. Our findings, in terms of various quantitative and qualitative performance indicators suggest that all the three strategies, either frontier-based or sampling-based, behave differently depending on the complexity of the environment and initial conditions, e.g., the entry point. There is certainly a trade-off among the various KPIs to select the best performing strategy for future UAV exploration missions. 

A hybrid strategy that combines the frontier-based and sampling-based approaches based on the complexity of the indoor environment and the limited resources may seem to be a promising direction for future research. Furthermore, we will test and validate the simulated exploration strategies on a DJI m210 UAV hardware platform in a real SAR indoor testbed environment. We also plan to investigate a fire-fighting relief target-oriented exploration scenario using an alternate perception sensor, i.e., a depth (RGB-D) camera, in our proposed framework under much more challenging environmental conditions.

%
%

\bibliographystyle{IEEEtran}
\bibliography{ref}

\end{document}